%% file: main.tex
\pdfoutput=1

\documentclass[11pt]{article}

\usepackage[preprint]{acl}

\usepackage{times}
\usepackage{latexsym}

\usepackage[T1]{fontenc}

\usepackage[utf8]{inputenc}

\usepackage{microtype}

\usepackage{inconsolata}

\usepackage{graphicx}
\usepackage{multirow}
\usepackage{multirow}
\usepackage{amsmath}
\usepackage{booktabs}
\usepackage{enumitem}
\usepackage{amsfonts}
\usepackage{makecell}
\usepackage{array}
\usepackage{geometry}
\usepackage{colortbl}

\newcommand{\ours}{DIVER}

\title{Beyond Quality: Unlocking Diversity in Ad Headline Generation with Large Language Models}

\author{
 \textbf{Chang Wang\textsuperscript{1}\thanks{~Equal Contribution.}},
 \textbf{Siyu Yan\textsuperscript{1,2}\footnotemark[1]\thanks{~Work done during an internship at Xiaohongshu Inc.}},
 \textbf{Depeng Yuan\textsuperscript{1}},
 \textbf{Yuqi Chen\textsuperscript{1}},
\\
 \textbf{Yanhua Huang\textsuperscript{1}\thanks{~Corresponding Author.}},
 \textbf{Yuanhang Zheng\textsuperscript{1}},
 \textbf{Shuhao Li\textsuperscript{1}},
 \textbf{Yinqi Zhang \textsuperscript{1}},
\\
 \textbf{Kedi Chen\textsuperscript{1,2}\footnotemark[2]},
 \textbf{Mingrui Zhu\textsuperscript{1}},
 \textbf{Ruiwen Xu\textsuperscript{1}}
\\
 \textsuperscript{1}Xiaohongshu Inc.,
 \textsuperscript{2}East China Normal University
\\
\texttt{\{wangchang2,yanhuahuang\}@xiaohongshu.com},
\texttt{yansiyu@stu.ecnu.edu.cn}
}
\begin{document}
\maketitle

\input{latex/0-abs}
\input{latex/1-intro}
\input{latex/2-works}
\input{latex/3-method}
\input{latex/4-exp}

\input{latex/5-conclusion}
\input{latex/6-limitations}
\input{latex/7-ethical}

\bibliography{custom}

\input{latex/appendix}

\end{document}

%% file: latex/0-abs.tex
\begin{abstract}


The generation of ad headlines plays a vital role in modern advertising, where both quality and diversity are essential to engage a broad range of audience segments.
Current approaches primarily optimize language models for headline quality or click-through rates (CTR), often overlooking the need for diversity and resulting in homogeneous outputs. 
To address this limitation, we propose {\ours}, a novel framework based on large language models (LLMs) that are jointly optimized for both diversity and quality. 
We first design a semantic- and stylistic-aware data generation pipeline that automatically produces high-quality training pairs with ad content and multiple diverse headlines. 
To achieve the goal of generating high-quality and diversified ad headlines within a single forward pass, we propose a multi-stage multi-objective optimization framework with supervised fine-tuning (SFT) and reinforcement learning (RL). 
Experiments on real-world industrial datasets demonstrate that {\ours} effectively balances quality and diversity. 
Deployed on a large-scale content-sharing platform serving hundreds of millions of users, our framework improves advertiser value (ADVV) and CTR by 4.0\% and 1.4\%.

\end{abstract}

%% file: latex/1-intro.tex
\section{Introduction}\label{sec:intro}

\begin{figure}[t!]
\centering
\includegraphics[width=\linewidth]{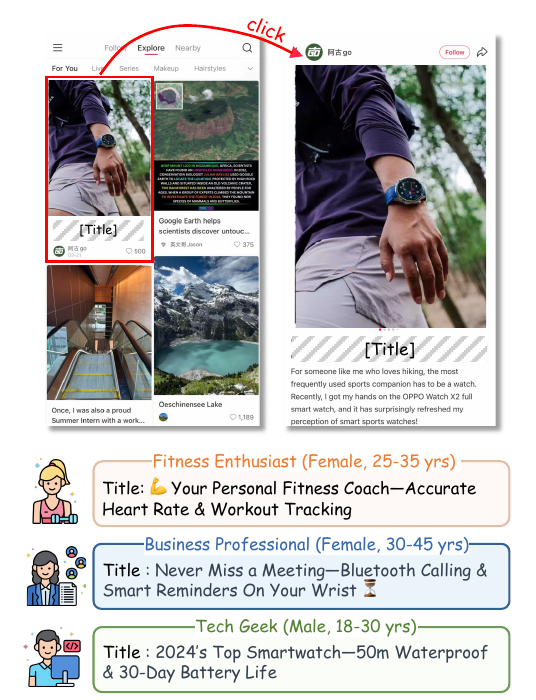}
\caption{An illustration of diversified ad headline generation in Xiaohongshu Inc., where fitness enthusiasts, business professionals, and tech geeks each receive relevant feature highlights in distinct styles.}
\label{fig:teaser}
\end{figure}

Ad headline generation plays an essential role in modern advertising, where the ability to produce diverse and engaging headlines directly influences campaign effectiveness~\cite{ao2021pens, zhang2022perception}. As shown in Figure~\ref{fig:teaser}, achieving this requires models that can flexibly adapt to different focal points, tones, and stylistic nuances.

Current approaches predominantly optimize for headline quality and click-through rate (CTR)~\cite{ao2023putyourvoice, song2023general}, often resulting in generic, one-size-fits-all outputs that fail to resonate with diverse audience segments. 
While recent advances in large language models (LLMs) have demonstrated strong generative capabilities~\cite{naveed2023comprehensive, achiam2023gpt, liu2024deepseek, huang2025towards}, applying them directly to ad headline generation introduces two key challenges.
First, although techniques like sampling-based~\cite{holtzman2020curiouscaseneuraltext, fan2018hierarchical} and constraint-based methods~\cite{lau2024dipperdiversitypromptsproducing} aim to enhance diversity, they often reduce robustness or limit adaptability. Moreover, fine-tuning LLMs struggles to balance diversity and quality~\cite{mai2024linguisticdiversity}, while separate models for each objective raise resource costs and hinder deployment.
Second, both SFT and RL typically rely on high-quality, task-specific datasets to achieve strong performance~\cite{ouyang2022traininglanguagemodelsfollow}. 
In ad headline generation, this requires diverse, high-quality headlines per content instance, the creation of which is labor-intensive.



To address these challenges, we propose {\ours}, a novel optimizing framework that reformulates diversified ad headline generation as a multi-stage, multi-objective optimization task. This framework enables the model to generate multiple diverse yet high-quality headlines in a single forward pass.
To achieve this goal, we first introduce a semantic and stylistic-aware data generation pipeline that automatically produces high-quality and diverse paired datasets. We then perform cold-start SFT on the synthetic data to equip the model with basic capabilities for generating multiple candidate headlines. Finally, we design a multi-objective reward function and apply reinforcement learning to optimize for quality and diversity explicitly.

Our main contributions are as follows:


\begin{itemize}
\item We propose {\ours}, a novel multi-stage multi-objective optimization framework that generates diverse, high-quality ad headlines.

\item We develop an automatic data generation pipeline that produces diverse, semantically and stylistically rich training examples.

\item We adopt a multi-stage training strategy with cold-start SFT and multi-objective RL to balance diversity and quality.

\item We deploy {\ours} on the Explore Feed of Xiaohongshu (a.k.a RedNote)\footnote{https://www.xiaohongshu.com/explore.}, a large-scale content-sharing platform, improving users' engagement and advertisers' satisfaction.
\end{itemize}


%% file: latex/2-works.tex
\section{Related Work}\label{sec:works}

\subsection{Ad Headline Generation}

Ad headline generation is a longstanding core task in natural language generation (NLG)~\cite{tevet2021evaluating}. 
Early methods relied on handcrafted templates, rule-based heuristics, or retrieval approaches~\cite{bartz2008natural, fujita2010automatic, thomaidou2013automated}, which produced generic and inflexible outputs. 
The emergence of neural models, particularly sequence-to-sequence and Transformer-based architectures~\cite{xu2019clickbait, kanungo2021ad,chen2025enhancing}, has substantially improved headline fluency and contextuality.
Despite these advances, most methods remain centered on optimizing headline quality and CTR, neglecting the importance of diversity in outputs.
To address the limitations of conventional approaches, recent research has explored personalization~\cite{ao2023putyourvoice, song2023general, tan2024offline} by incorporating user preferences or contextual signals to tailor outputs to individual users. 
However, these personalized methods often focus on specific audiences without systematically improving headline diversity. Furthermore, the absence of multi-reference datasets continues to hinder the creation of varied ad content.
To address these limitations, we propose a multi-stage, multi-objective framework with automatic data generation to systematically enhance headline diversity.

\subsection{LLMs for Diversity}


Recent advances in LLMs~\cite{naveed2023comprehensive, achiam2023gpt, liu2024deepseek} have significantly enhanced automatic text generation across a wide range of tasks, from headline generation~\cite{lian2025panoramic} to more open-ended creative writing and content creation~\cite{mai2024linguisticdiversity}.
To encourage diversity in generated texts, researchers have explored various stochastic decoding strategies~\cite{holtzman2020curiouscaseneuraltext, fan2018hierarchical} as well as prompt engineering techniques~\cite{lau2024dipperdiversitypromptsproducing}. However, while stochastic decoding can increase diversity, it often leads to uncontrollable outputs with compromised text quality and coherence. On the other hand, prompt engineering typically depends on pre-defined labels or templates, which inherently limit the flexibility and generalization of the models to new domains or tasks.
More recently, researchers have begun investigating diversity-driven training objectives~\cite{mai2024linguisticdiversity} to explicitly promote diversity during training, but the trade-off between quality and diversity remains underexplored.
Although methods such as SFT and RL on task-specific datasets can improve headline quality~\cite{mai2024linguisticdiversity}, they often produce deterministic outputs by overfitting to dominant patterns~\cite{kirk2024understandingeffectsrlhfllm}, limiting diversity. To address these issues, our solution combines synthetic data and multi-objective RL to jointly optimize diversity, quality, and CTR, generating high-quality headlines in a single pass.

%% file: latex/3-method.tex
\begin{figure*}[!t]
\centering
\includegraphics[width=\linewidth]{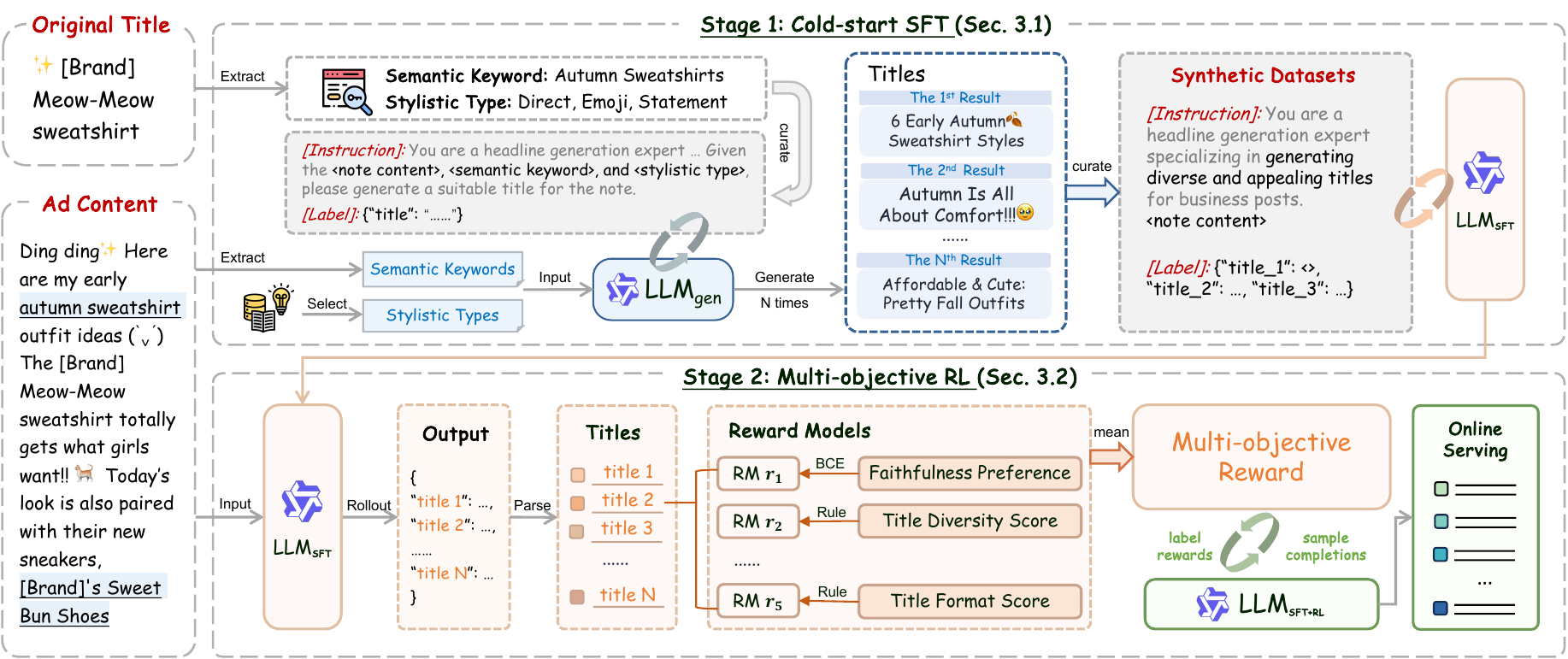}
\caption{Overview of the {\ours} framework. Our approach first performs synthetic data-augmented SFT to enable basic diversity in headline generation. This is followed by multi-objective RL to further enhance the diversity and quality of generated headlines through a composite reward function.}
\label{fig:framework}
\end{figure*}

\section{Method}\label{sec:method}

We introduce {\ours} for generating diverse ad headlines, as illustrated in Figure~\ref{fig:framework}. {\ours} employs a multi-stage multi-objective training pipeline consisting of (1) synthetic data-augmented fine-tuning for cold-start SFT (Section~\ref{sec:coldstart}), and (2) multi-objective reinforcement learning for enhancing quality and diversity (Section~\ref{sec:rl}).

\subsection{Synthetic Data for Cold-start SFT}\label{sec:coldstart}

Creating datasets with multiple diverse headlines for each ad content is labor-intensive. To address this, we propose a data generation pipeline that leverages LLMs to synthesize training samples for cold-start supervised fine-tuning. 

\paragraph{Semantic- and Stylistic-aware Data Enrichment.}

Given an industrial dataset $\mathcal{D}$ consisting of original headlines and their corresponding ad content, we first employ open source LLMs to annotate each headline with its semantic keyword and stylistic type\footnote{Throughout the paper, we define an ad headline style along three dimensions: directness (direct vs. indirect), emoji usage (with emoji vs. without emoji), and rhetorical type (question, exaggeration, metaphor, or statement). Combining these dimensions yields a total of 16 distinct headline styles.}, resulting in a dataset $\mathcal{D}'$ composed of quadruples in the format $\langle$ad content, semantic keyword, stylistic type, headline$\rangle$. We then fine-tune a generator $\pi_{\theta_{\text{gen}}}$ with $\langle$ad content, semantic keyword, stylistic type$\rangle$ as input and headline as the target label, enabling it to generate headlines conditioned on both semantic and stylistic cues. 

\paragraph{Controlled Diverse Headline Generation.}

For each ad content, we prompt an LLM to generate multiple semantically distinct keywords conditioned on the ad content, each paired with a randomly selected stylistic type. These semantic keywords and style pairs, combined with the ad content, are fed into $\pi_{\theta_{\text{gen}}}$ to produce diverse headline sets in both meaning and tone. 
Finally, we further perform an LLM-based verification step to ensure that each generated headline covers the required semantic keywords and matches the assigned stylistic type. Only headlines that pass this verification are retained for subsequent training.

\paragraph{Dataset Construction and Training.}

The synthetic dataset consists of ad content as input and a set of multiple headlines as output, structured in a consistent template format, as illustrated in Figure~\ref{fig:framework}.
During cold-start SFT, we input ad content into $\pi_{\theta_{\text{sft}}}$ and train it to generate multiple diverse headlines in a structured format, allowing the model to produce semantically and stylistically varied outputs in a single pass.


\subsection{Multi-objective Reinforcement Learning}\label{sec:rl}


While SFT with synthetic data can encourage basic diversity, supervised learning alone often leads to repetitive outputs and mediocre phrasing~\cite{kirk2024understandingeffectsrlhfllm}. To overcome this, we adopt multi-objective reinforcement learning with a tailored reward function, a widely used approach for balancing and optimizing multiple competing objectives in RLHF~\cite{wu2023fine, dai2024saferlhf}.

\subsubsection{Reward Design}

We design fine-grained reward functions to guide the model in generating diverse, faithful, and engaging headlines, with the overall reward averaged across five components. Further details on the reward function design and reward model training are provided in Appendix~\ref{apdx:reward}.

\paragraph{Diversity Reward.}

This reward combines semantic and stylistic diversity. Semantic diversity is measured as the complement of the average pairwise BLEU score~\cite{papineni2002bleu}, while stylistic diversity reflects the coverage of predefined style types. The overall reward is computed as:
\[
r_{\text{diversity}} = \frac{1 - \text{Pair-BLEU}(Y) + \text{Coverage}(Y)}{2} ~,
\]
where $Y = \{y_1, \ldots, y_N\}$ is the set of generated headlines.

\paragraph{Quality Reward.}

We evaluate the quality of each headline in terms of faithfulness to the input document, using a fine-tuned model that outputs a faithfulness score between $0$ and $1$. The reward reflects the proportion of headlines that meet or exceed a given faithfulness threshold.

\paragraph{CTR Reward.}

To reflect user satisfaction, we use a CTR prediction model trained on historical user interaction logs to score each generated headline. The user preference reward is the average CTR score across all generated headlines.

\paragraph{Quantity Reward.}

This reward encourages the model to output the predefined number of headlines by explicitly comparing the actual count with the specified target number. 
The reward grows linearly with the number of generated headlines and saturates when the target number is reached.

\paragraph{Format Reward.}

This reward is higher if the model is able to generate the headlines in a correct and easily parsed JSON format, making it straightforward and efficient to extract each headline.

\subsubsection{RL Optimization}

We formulate headline generation as a policy learning task, where the model $\pi_{\theta}$ generates $N$ headlines per content $x$ in a single pass, producing outputs $Y = \{y_1, \ldots, y_N\}$. During RL optimization, we repeatedly sample headline sets, compute the composite reward, and update the model using the GRPO algorithm~\cite{shao2024deepseekmath}:
\begin{equation*}
    \max_{\theta} \ \mathbb{E}_{x\sim D,\, Y\sim \pi_\theta(\cdot|x)} \left[R(x, Y)\right] - \beta D_{\mathrm{KL}}\left(\pi_{\theta} \| \pi_{\theta_{\text{sft}}}\right),
\end{equation*}
where $R(x, Y)$ denotes the composite reward for the generated set $Y$, $\beta$ controls the strength of the KL penalty, and $\pi_{\theta_{\text{sft}}}$ is the reference model.

During inference, each input advertising content is passed through $\pi_{\theta}$ to generate a set of headlines for online serving.

%% file: latex/4-exp.tex
\section{Experiments}\label{sec:exp}

\subsection{Experimental Setup}\label{sec:exp_details}


\begin{table*}[ht]
\centering
\definecolor{graybg}{gray}{0.92}
\resizebox{\linewidth}{!}{
\begin{tabular}{lccccccccc}
\toprule
\multirowcell{2}[-3pt][c]{Method}
  & \multicolumn{5}{c}{\textbf{Diversity}} 
  & \multicolumn{4}{c}{\textbf{Quality}} \\
\cmidrule(lr){2-6} \cmidrule(lr){7-10}
 & PairBLEU $\downarrow$ & SelfBLEU $\downarrow$ & DisNGram $\uparrow$ 
 & CosSim $\downarrow$ & StyleCov $\uparrow$ 
 & NLI $\uparrow$ & Rouge-1 $\uparrow$ & Rouge-2 $\uparrow$ & Rouge-L $\uparrow$ \\
\midrule

\multicolumn{10}{l}{\textit{Base: Closed-source Models}} \\
GPT-4o            & 10.46 & \underline{40.97} & 47.39 & 50.57 & 50.73\% & 70.48 & 16.19 & 4.71 & 9.91 \\
Claude-3.5        & 8.21 & 43.89 & \textbf{53.02} & 47.46 & 45.63\% & 75.73 & 14.29 & 3.89 & 8.69 \\
\midrule

\multicolumn{10}{l}{\textit{Base: Open-source Models}} \\
Qwen2.5-72B       & 21.41 & 55.02 & 47.62 & 78.00 & 39.26\% & 72.95 & \textbf{17.93} & 6.05 & 10.87 \\
DeepSeek V3       & 20.91 & 53.37 & 43.50 & 54.88 & 42.78\% & \textbf{83.83} & 17.14 & 5.29 & 10.64 \\
\midrule

\multicolumn{10}{l}{\textit{Base: Qwen2.5-14B-Instruct}} \\
PEFT                         & \underline{5.71}  & 47.89 & 38.43 & \underline{42.16} & \underline{60.20}\% & 75.65 & \underline{17.28} & \underline{6.93} & \textbf{11.22} \\
\rowcolor{graybg}
\textbf{{\ours}}              & \textbf{2.08}  & \textbf{35.93} & \underline{52.92} & \textbf{28.93} & \textbf{63.42}\% & \underline{76.72} & 16.71 & \textbf{7.30} & \underline{10.91} \\
\bottomrule
\end{tabular}
}
\caption{Performance comparison of baseline models and our method. The best value in each column is \textbf{bolded}, the second best is \underline{underlined}. Row with a gray background stand for our method.}
\label{exp:baseline}
\end{table*}

\begin{table*}[ht]
\centering
\definecolor{graybg}{gray}{0.92}
\resizebox{\linewidth}{!}{
\begin{tabular}{lccccccccc}
\toprule
\multirowcell{2}[-3pt][c]{Method} 
  & \multicolumn{5}{c}{\textbf{Diversity}} 
  & \multicolumn{4}{c}{\textbf{Quality}} \\
\cmidrule(lr){2-6} \cmidrule(lr){7-10}
 & PairBLEU $\downarrow$ & SelfBLEU $\downarrow$ & DisNGram $\uparrow$ 
 & CosSim $\downarrow$ & StyleCov $\uparrow$ 
 & NLI $\uparrow$ & Rouge-1 $\uparrow$ & Rouge-2 $\uparrow$ & Rouge-L $\uparrow$ \\
\midrule
\textbf{{\ours}}              & 2.08  & 35.93 & 52.92 & 28.93 & 63.42\% & 76.72 & 16.71 & 7.30 & 10.91 \\
\midrule
\multicolumn{10}{l}{\textit{Ablation Study: Components}} \\
\hspace{1em}~w/o Data & 
6.84\textsubscript{\textcolor{red}{$\uparrow4.76$}} & 
45.47\textsubscript{\textcolor{red}{$\uparrow9.54$}} & 
44.65\textsubscript{\textcolor{red}{$\downarrow8.27$}} & 
42.64\textsubscript{\textcolor{red}{$\uparrow13.71$}} & 
58.49\%\textsubscript{\textcolor{red}{$\downarrow4.93$}} & 
70.60\textsubscript{\textcolor{red}{$\downarrow6.12$}} & 
16.15\textsubscript{\textcolor{red}{$\downarrow0.56$}} & 
6.32\textsubscript{\textcolor{red}{$\downarrow0.98$}} & 
10.50\textsubscript{\textcolor{red}{$\downarrow0.41$}} \\

\hspace{1em}~w/o RL & 
7.82\textsubscript{\textcolor{red}{$\uparrow5.74$}} & 
48.81\textsubscript{\textcolor{red}{$\uparrow12.88$}} & 
48.15\textsubscript{\textcolor{red}{$\downarrow4.77$}} & 
40.01\textsubscript{\textcolor{red}{$\uparrow11.08$}} & 
57.04\%\textsubscript{\textcolor{red}{$\downarrow6.38$}} & 
73.24\textsubscript{\textcolor{red}{$\downarrow3.48$}} & 
18.47\textsubscript{\textcolor{teal}{$\uparrow1.76$}} & 
8.94\textsubscript{\textcolor{teal}{$\uparrow1.64$}} & 
12.49\textsubscript{\textcolor{teal}{$\uparrow1.58$}} \\

\hspace{1em}~w/o Both & 
10.12\textsubscript{\textcolor{red}{$\uparrow8.04$}} & 
50.20\textsubscript{\textcolor{red}{$\uparrow14.27$}} & 
45.18\textsubscript{\textcolor{red}{$\downarrow7.74$}} & 
47.75\textsubscript{\textcolor{red}{$\uparrow18.82$}} & 
53.35\%\textsubscript{\textcolor{red}{$\downarrow10.07$}} & 
67.59\textsubscript{\textcolor{red}{$\downarrow9.13$}} & 
16.86\textsubscript{\textcolor{teal}{$\uparrow0.15$}} & 
7.04\textsubscript{\textcolor{red}{$\downarrow0.26$}} & 
11.13\textsubscript{\textcolor{teal}{$\uparrow0.22$}} \\
\midrule

\multicolumn{10}{l}{\textit{Ablation Study: Reward Functions}} \\
\hspace{1em}~w/o Diversity & 
4.88\textsubscript{\textcolor{red}{$\uparrow2.80$}} & 
48.21\textsubscript{\textcolor{red}{$\uparrow12.28$}} & 
49.17\textsubscript{\textcolor{red}{$\downarrow3.75$}} & 
32.98\textsubscript{\textcolor{red}{$\uparrow4.05$}} & 
49.40\%\textsubscript{\textcolor{red}{$\downarrow14.02$}} & 
76.99\textsubscript{\textcolor{teal}{$\uparrow0.27$}} & 
15.72\textsubscript{\textcolor{red}{$\downarrow0.99$}} & 
6.67\textsubscript{\textcolor{red}{$\downarrow0.63$}} & 
10.45\textsubscript{\textcolor{red}{$\downarrow0.46$}} \\

\hspace{1em}~w/o Quality & 
0.30\textsubscript{\textcolor{teal}{$\downarrow1.78$}} & 
16.78\textsubscript{\textcolor{teal}{$\downarrow19.15$}} & 
61.29\textsubscript{\textcolor{teal}{$\uparrow8.37$}} & 
30.79\textsubscript{\textcolor{red}{$\uparrow1.86$}} & 
39.44\%\textsubscript{\textcolor{red}{$\downarrow23.98$}} & 
75.09\textsubscript{\textcolor{red}{$\downarrow1.63$}} & 
12.36\textsubscript{\textcolor{red}{$\downarrow4.35$}} & 
5.03\textsubscript{\textcolor{red}{$\downarrow2.27$}} & 
8.30\textsubscript{\textcolor{red}{$\downarrow2.61$}} \\

\hspace{1em}~w/o CTR & 
3.10\textsubscript{\textcolor{red}{$\uparrow1.02$}} & 
41.21\textsubscript{\textcolor{red}{$\uparrow5.28$}} & 
52.82\textsubscript{\textcolor{red}{$\downarrow0.10$}} & 
31.67\textsubscript{\textcolor{red}{$\uparrow2.74$}} & 
46.69\%\textsubscript{\textcolor{red}{$\downarrow16.73$}} & 
76.20\textsubscript{\textcolor{red}{$\downarrow0.52$}} & 
17.03\textsubscript{\textcolor{teal}{$\uparrow0.32$}} & 
8.05\textsubscript{\textcolor{teal}{$\uparrow0.75$}} & 
11.48\textsubscript{\textcolor{teal}{$\uparrow0.57$}} \\

\hspace{1em}~w/o Quantity & 
1.69\textsubscript{\textcolor{teal}{$\downarrow0.39$}} & 
15.29\textsubscript{\textcolor{teal}{$\downarrow20.64$}} & 
84.06\textsubscript{\textcolor{teal}{$\uparrow31.14$}} & 
27.65\textsubscript{\textcolor{teal}{$\downarrow1.28$}} & 
48.73\%\textsubscript{\textcolor{red}{$\downarrow14.69$}} & 
76.34\textsubscript{\textcolor{red}{$\downarrow0.38$}} & 
11.11\textsubscript{\textcolor{red}{$\downarrow5.60$}} & 
3.45\textsubscript{\textcolor{red}{$\downarrow3.85$}} & 
7.28\textsubscript{\textcolor{red}{$\downarrow3.63$}} \\

\hspace{1em}~w/o Format & 
3.05\textsubscript{\textcolor{red}{$\uparrow0.97$}} & 
41.08\textsubscript{\textcolor{red}{$\uparrow5.15$}} & 
41.10\textsubscript{\textcolor{red}{$\downarrow11.82$}} & 
30.16\textsubscript{\textcolor{red}{$\uparrow1.23$}} & 
56.57\%\textsubscript{\textcolor{red}{$\downarrow6.85$}} & 
76.14\textsubscript{\textcolor{red}{$\downarrow0.58$}} & 
15.23\textsubscript{\textcolor{red}{$\downarrow1.48$}} & 
6.49\textsubscript{\textcolor{red}{$\downarrow0.81$}} & 
10.15\textsubscript{\textcolor{red}{$\downarrow0.76$}} \\
\bottomrule
\end{tabular}
}
\caption{Ablation study of {\ours}. Subscripts show differences compared with {\ours}, with red indicating worse and green indicating better performance.}
\label{tab:ablation}
\end{table*}

\paragraph{Datasets.}

To our knowledge, no publicly available large-scale dataset exists specifically for the advertising domain. Therefore, we constructed an industrial dataset by collecting commercial ad logs from a leading content-sharing platform for both training and evaluation. Further details regarding dataset statistics, construction, and preprocessing can be found in Appendix~\ref{apdx:data_processing}.

\paragraph{Baselines.}

We chose Qwen2.5-14B-Instruct as a base model~\cite{qwen2025qwen25technicalreport} to conduct SFT and RL training and generate multiple ad headlines. Additional experimental settings are provided in Appendix~\ref{apdx:detailed-exp-setup}.  To comprehensively evaluate our approach, we compared it against two categories of baselines. First, we include state-of-the-art open-source and proprietary models, such as GPT-4o~\cite{openai2024gpt4ocard}, Claude-3.5-Sonnet~\cite{anthropic2024claude3.5}, DeepSeek-V3~\cite{deepseekai2025deepseekv3technicalreport}, and Qwen2.5-72B-Instruct~\cite{qwen2025qwen25technicalreport}. Second, we consider fine-tuning-based methods, including Possibility Exploration Fine-Tuning (PEFT)~\cite{mai2024linguisticdiversity}, applied to our generated dataset. All models were tested on the same datasets under controlled settings.


\paragraph{Evaluation Metrics.}

We adopt a dual-aspect evaluation framework that considers both diversity and quality. To assess diversity, we measure both lexical and semantic variation among generated titles using Pairwise BLEU~\cite{papineni2002bleu}, Self-BLEU~\cite{zhu2018texygen}, Distinct N-Gram~\cite{li2015diversity}, and Cosine Similarity~\cite{slaton1986retrieval}\footnote{CosSim is computed using Sentence-BERT at: \url{https://huggingface.co/uer/sbert-base-chinese-nli}}. We evaluate style diversity via Style Coverages. 
For quality, we assess both faithfulness and content relevance of the headlines using NLI-based evaluation~\cite{yoran2023making}\footnote{NLI-based evaluation is performed with mDeBERTa-v3-base at: \url{https://huggingface.co/MoritzLaurer/mDeBERTa-v3-base-xnli-multilingual-nli-2mil7}.}, Rouge-1, Rouge-2, and Rouge-L~\cite{chin2004rouge}.

\subsection{Main Results}


As shown in Table~\ref{exp:baseline}, {\ours} demonstrates superior performance over other methods across most diversity metrics. Specifically, it achieves the lowest scores for both Pairwise-BLEU and Self-BLEU, indicating minimal redundancy among generated titles, and covers the largest proportion of target styles. Meanwhile, our approach maintains a high quality score that is on par with advanced baselines such as Claude-3.5-Sonnet and DeepSeek V3. Compared to prompting and fine-tuning strategies, {\ours} consistently produces ad headlines that are more diverse and stylistically rich while remaining faithful to the original content. These findings underscore the capability of our approach to produce ad headlines that balance diversity and quality.

\subsection{Ablation Studies}

\begin{table*}[ht]
\centering
\small
\begin{tabular}{
    p{2.4cm}
    p{4cm}
    p{4cm}
    p{4cm}
}
\toprule
\textbf{Attribute} & \textbf{Wedding Suit Ad} & \textbf{Home Improvement Fence Ad} & \textbf{Anti-Aging Injection Ad} \\
\midrule
\textbf{Original Title} & My husband in a black suit was surrounded by onlookers at our wedding 
& So easy, you'll get it at a glance! Outdoor Wood-Plastic Fence Installation Tutorial
& 30+ Anti-Aging Injections | Don't Ignore Perioral Aging \\
\midrule
\textbf{Ad Content Summary} 
& A black suit with a white shirt stole the show at the wedding; many guests were impressed by the sharp look and classic style. Tips: choose quality fabric and tailoring, and pair with classic accessories.
& Villa's wood-plastic fence was installed in one day, thanks to its simple design and skilled craftsmanship. DIY encouraged, showing style and benefits (teak and black aluminum, suitable for home improvement).
& Details on facial aging (apple cheeks, nasolabial folds, marionette lines) and advanced injectable anti-aging techniques. Focus on individualized, balanced correction for youthful appearance. \\
\midrule
\textbf{User Type 1} 
& Male
& DIY \& fitness lover
& Women of suitable age \\
\textbf{Generated Title 1} 
& As expected! Black suit with white shirt—unbeatable classic combo.
& DIY home improvement + Get a workout! Experience the joy of hands-on installation.
& Smart injectable anti-aging: Don't ignore mouth area rejuvenation. \\
\midrule
\textbf{User Type 2}
& Female
& DIY \& aesthetics lover
& Young people \\
\textbf{Generated Title 2}
& Thank you for the custom suit! The groom looked so handsome.
& Beautiful teak panels, stunning effect—upgrade your yard effortlessly!
& [Tips] Prevent ``Sagging Apple Cheeks''---start early! \\
\bottomrule
\end{tabular}
\caption{Examples of generating different ad titles for different users based on ad content. Each column is an ad type; each row gives a corresponding attribute or personalized title.}
\label{tab:ad_title_generation}
\end{table*}

\begin{table}[t]
\centering
\resizebox{\linewidth}{!}{
\begin{tabular}{lcccc}
\toprule
\textbf{Model} &\textbf{ADVV} & \textbf{CTR} & \textbf{IMP} & \textbf{CPM}    \\
\midrule
Sampling + SFT    & +2.2\%  & +0.7\% & +1.3\% & +1.2\%  \\
\textbf{{\ours}} & +4.0\%  & +1.4\% & +2.4\% & +2.0\%  \\
\bottomrule
\end{tabular}
}
\caption{Online A/B test results comparing different models using advertiser values (ADVV), click-through rate (CTR), impression (IMP), and cost per mile (CPM).}
\label{tab:abtest}
\end{table}



To evaluate the contribution of each component in our framework, we perform ablation studies by selectively removing the semantic- and stylistic-aware data augmentation pipeline (w/o Data), the multi-objective reinforcement learning phase (w/o RL), or both (w/o Both). As shown in Table~\ref{tab:ablation}, removing either the augmented data or the RL stage leads to noticeable declines in both diversity and faithfulness. 
Excluding both components leads to the weakest performance, while {\ours} achieves the best balance of diversity and quality, with the highest quality score, broadest style coverage, and lowest redundancy, highlighting the value of data augmentation and RL optimization.

\subsection{Analysis of Multi-objective RL}

We further analyze the effectiveness of each reward function within the multi-objective RL.
As shown in Table~\ref{tab:ablation}, removing the diversity reward (w/o Diversity) leads to a significant decrease in all diversity metrics,
while minimal improvement in headline quality, indicating its key role in promoting output variety. Removing the quality reward (w/o Quality) improves diversity but sharply reduces faithfulness and informativeness, highlighting the quality signal’s importance.
Removing CTR, quantity, or format rewards leads to declines in style coverage, diversity, or overall performance, indicating that all components are vital for balancing diversity and quality.







\subsection{Online Case Study}

Table~\ref{tab:ad_title_generation} shows diverse ad headlines generated by {\ours}. For each advertisement, the model generates multiple candidate titles that cover different expressions or emphases, reflecting varied user perspectives (e.g., male or female) and interests (e.g., functional vs. aesthetic appeal, practical tips vs. emotional resonance). This demonstrates its ability to produce a wide range of high-quality and diverse ad headlines for online personalization.   


\subsection{Online A/B Test}

We have deployed {\ours} on the Explore Feed of Xiaohongshu, a large-scale content sharing platform, where advertising performance is primarily measured by advertiser value (ADVV)~\cite{chai2025longer,timmaraju2023towards} and click-through rate (CTR). During online serving, we first generate $30$ ad headlines with {\ours}. To enable personalization, we select the headline most semantically similar to the user profile. Online A/B testing results, as shown in Table~\ref{tab:abtest}, demonstrate the practical effectiveness of {\ours}. Specifically, models using high-temperature sampling and SFT without synthetic data achieve moderate improvements over the base model in both ADVV (+2.2\%) and CTR (+0.7\%). {\ours}, combining synthetic data, cold-start SFT, and multi-objective RL, achieves a further boost, with ADVV increasing by 4.0\% and CTR by 1.4\%. These results show that our approach enhances both headline quality and diversity while delivering business impact.


%% file: latex/5-conclusion.tex
\section{Conclusion}\label{sec:conclusion}

This paper addresses the challenge of generating ad headlines that are both high-quality and diverse, which is crucial for attracting and engaging various user segments. By introducing a semantic- and stylistic-aware data generation pipeline and a multi-stage, multi-objective optimization framework combining SFT and RL, our method effectively balances diversity and quality. We have successfully deployed {\ours} on a large-scale content-sharing platform, achieving significant gains in core metrics for the advertising system.

%% file: latex/6-limitations.tex
\section*{Limitations}

Although {\ours} performs well in generating diverse, high-quality ad headlines, several limitations remain. Synthetic data may introduce noise or stylistic bias, limiting personalization and generalization. Diversity in long-tail categories suffers from data scarcity, and fixed reward metrics may overlook nuanced user preferences. Deployment also faces challenges in latency, scalability, and adapting to user trends. Future work will focus on enriching long-tail data, incorporating richer signals, and adopting more adaptive rewards to improve practical effectiveness.

%% file: latex/7-ethical.tex
\section*{Ethical Considerations}

All datasets used in this study used are properly licensed and contain no private or sensitive user information. Generated ad headlines require advertiser approval before use, and we apply rigorous post-processing, including quality control and risk assessment, prior to online deployment. An online blacklist system further ensures rapid removal of any problematic content. These measures collectively safeguard user privacy, content integrity, and platform safety throughout our framework.

%% file: latex/appendix.tex
\clearpage
\appendix

\section{Detailed Reward Design} \label{apdx:reward}

This section details rule-based (diversity, quantity, format) and model-based (quality, CTR) rewards.


\paragraph{Details about Rule-based Rewards.}

We formulate diversity, quantity, and format rewards as rule-based rewards. For the diversity reward, semantic diversity is computed as the average pairwise BLEU score within the generated set, i.e,
\[
\text{Pair-BLEU}(Y) = \frac{1}{Z} \sum_{i=1}^N \sum_{\substack{j=1 \\ i \ne j}}^N \text{BLEU}(y_i, y_j) ~,
\]
where $Z=N \cdot (N-1)$. Style diversity is measured by the proportion of distinct style categories presents in the generated headlines, where we prompt DeepSeek-V3 to classify the style of the headline. The quantity reward encourages generating at least $T$ headlines, defined as $r_{\text{quantity}} = \min(1, N / T)$. The format reward is 1 if the output is valid JSON; otherwise, it is 0.

\paragraph{Details about Quality Reward.}

To promote high-quality headline generation, we use a human-labeled quality reward. Headlines sampled via high-temperature SFT are labeled as high-quality ($1$) or not ($0$) and used to train a binary classifier $f_{\text{quality}}(\cdot)$ with content and headline as input, optimized using binary cross-entropy. The quality reward during RL is the average predicted score across all headline-content pairs.


\paragraph{Details about CTR Reward.}

To optimize user satisfaction, we train a CTR-based reward model using online interaction logs. For each 10,000 notes, multiple headlines are generated via high-temperature SFT, and user interaction data is used to label the top and bottom third of headlines by CTR as positive and negative samples. This yields 40,000 headline pairs to train a CTR prediction model $f_{\text{CTR}}(h, x)$ with headline $h$ and content $x$ as input and optimize with a pairwise margin loss:
\[
\mathcal{L} = \frac{1}{N} \sum_{i=1}^{N} \max(0, 0.3 - s_i^+ + s_i^-) ~,
\]
where $N$ represents the batch size, $s_i^{+}=f_{\text{CTR}}(h_i^{+}, x_i)$ and $h_i^{+}$ is the positive headline for the content $x_i$. $s_i^{-}=f_{\text{CTR}}(h_i^{-}, x_i)$ represents the predicted score for the negative headline.
All user data is anonymized. During RL training, the CTR reward is the average predicted score across all generated headline and content pairs.



\section{Dataset Construction and Processing}\label{apdx:data_processing}


This section introduces the dataset used for our ad headline generation task.

\paragraph{Raw Data.} 
Our dataset comprises commercial ad notes from a major content-sharing platform in China. 
To ensure privacy and compliance, all personal information was anonymized. Each instance includes the original title, content, topics, caption, and taxonomy, offering key semantic and stylistic cues for headline generation.

\paragraph{Preprocessing.} 
To ensure data quality and representativeness, we prioritized titles with high CTR while filtering out those with inflated CTR due to excessive exposure. We also balanced category distribution, removed duplicate or near-duplicate ads based on string similarity, and cleaned records with missing fields, repetition, or encoding errors.

\paragraph{Split and Statistics.} 
The dataset was chronologically split into training and test sets to reflect real-world usage. Table~\ref{tab:stats} presents key statistics. This dataset provides a high-quality benchmark for training and evaluating ad headline generation models.

\begin{table}[htbp]
    \centering
    \begin{tabular}{lrr}
        \toprule
        \textbf{Subset}      & \textbf{Number of Instances} \\
        \midrule
        SFT training set     & 50,000                       \\
        RL training set      & 79,334                       \\
        Test set             & 3,000                        \\
        \bottomrule
    \end{tabular}
    \caption{Dataset Statistics}
    \label{tab:stats}
\end{table}

\section{Experimental Setups}
\label{apdx:detailed-exp-setup}

We selected Qwen2.5-14B~\cite{qwen2025qwen25technicalreport} as the base model for experiments, and conducted training on a single server with 8 NVIDIA H800 GPUs.

\paragraph{Supervised Fine-tuning.}

The model was fine-tuned for 3 epochs on 50,000 samples, using a maximum input length of 6,000 tokens, a learning rate $1 \times 10^{-5}$, and bf16 precision.

\paragraph{Reinforcement Learning.}

We used the GRPO algorithm~\cite{shao2024deepseekmath} with full-parameter fine-tuning. RL training was performed on 79,334 samples, with an input cutoff of 4,096 tokens, a learning rate of $3 \times 10^{-6}$, and bf16 precision.



\section{Detailed Prompts}
\label{apdx:prompts}

\begin{figure*}[!t]
    \centering
    \includegraphics[width=\linewidth]{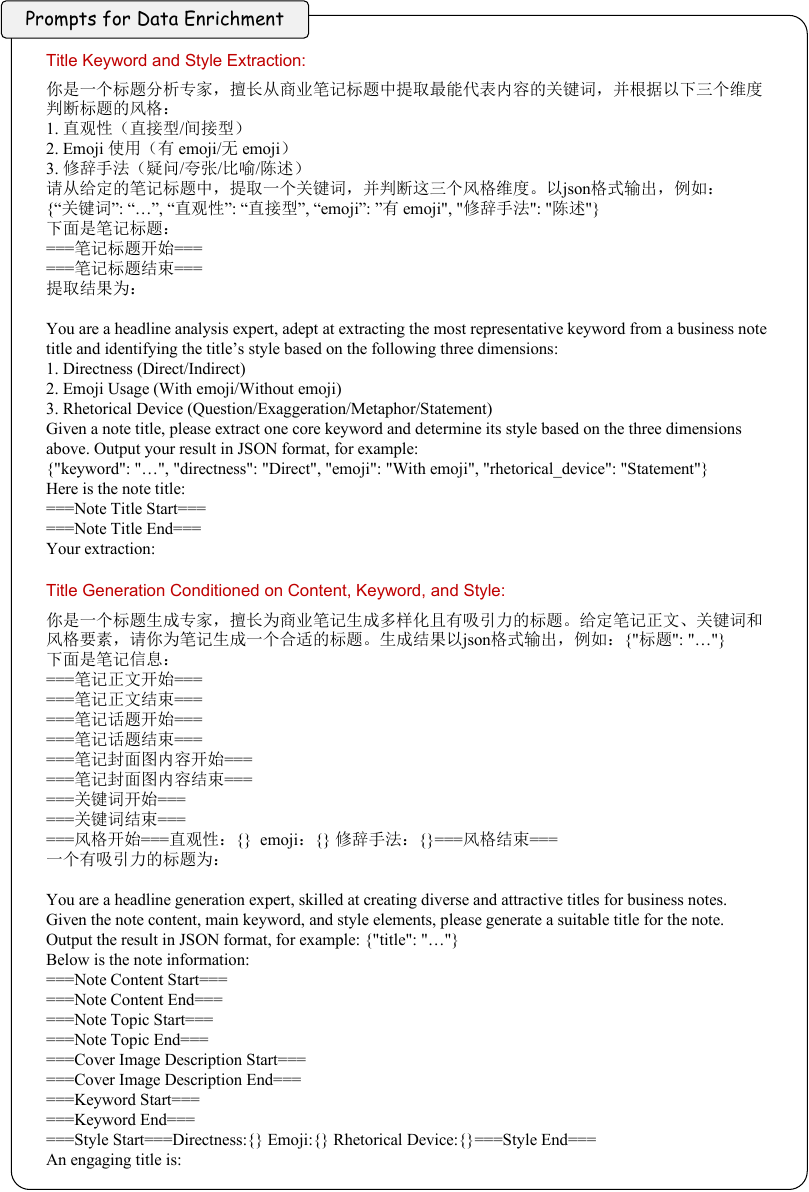}
    \caption{
    Prompts for the data enrichment.}
    \label{fig:prompts_data}
\end{figure*}

\begin{figure*}[!t]
    \centering
    \includegraphics[width=\linewidth]{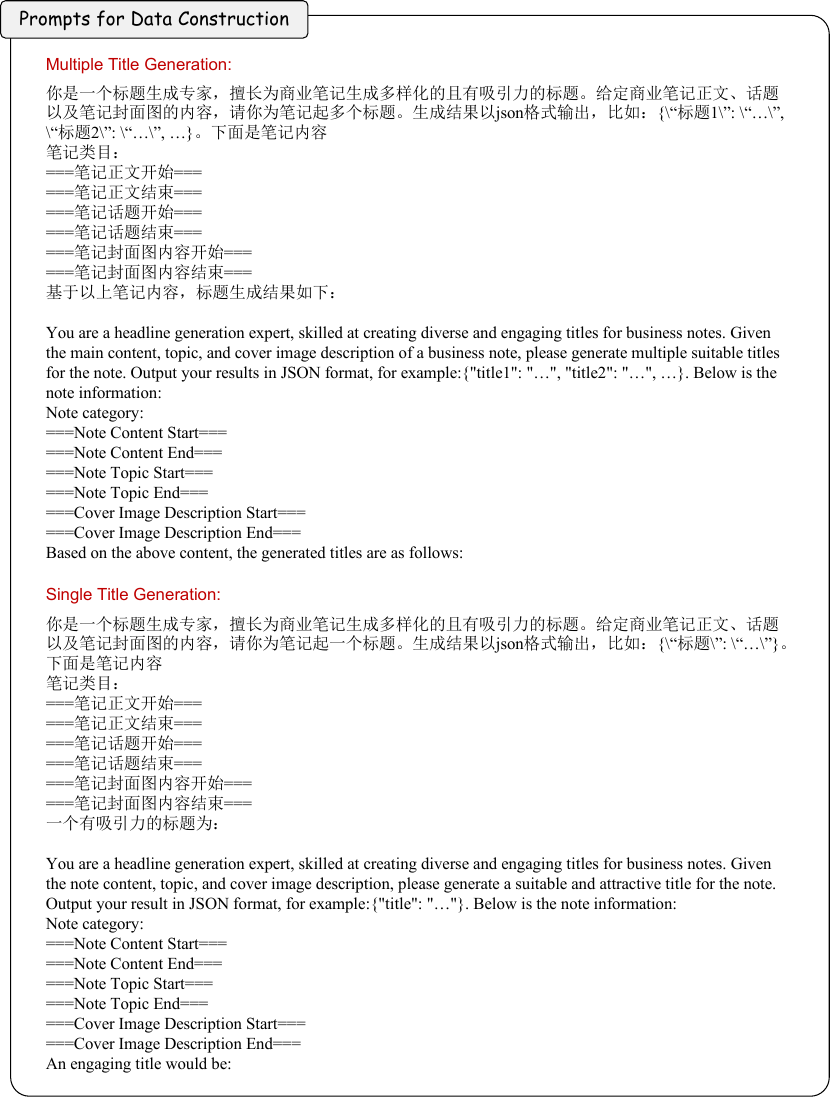}
    \caption{
    Prompts for the data construction.}
    \label{fig:prompts_train}
\end{figure*}

The key prompts used for data enrichment and data construction are shown in Figure~\ref{fig:prompts_data} and Figure~\ref{fig:prompts_train}.

%% file: main.bbl
\begin{thebibliography}{38}
\providecommand{\natexlab}[1]{#1}

\bibitem[{Achiam et~al.(2023)Achiam, Adler, Agarwal, Ahmad, Akkaya, Aleman, Almeida, Altenschmidt, Altman, Anadkat et~al.}]{achiam2023gpt}
Josh Achiam, Steven Adler, Sandhini Agarwal, Lama Ahmad, Ilge Akkaya, Florencia~Leoni Aleman, Diogo Almeida, Janko Altenschmidt, Sam Altman, Shyamal Anadkat, et~al. 2023.
\newblock Gpt-4 technical report.
\newblock \emph{arXiv preprint arXiv:2303.08774}.

\bibitem[{Anthropic(2024)}]{anthropic2024claude3.5}
Anthropic. 2024.
\newblock \href {https://www-cdn.anthropic.com/fed9cc193a14b84131812372d8d5857f8f304c52/Model_Card_Claude_3.pdf} {Claude 3.5 sonnet model card addendum}.

\bibitem[{Ao et~al.(2023)Ao, Luo, Wang, Yang, Chen, Qiao, He, and Xie}]{ao2023putyourvoice}
Xiang Ao, Ling Luo, Xiting Wang, Zhao Yang, Jiun-Hung Chen, Ying Qiao, Qing He, and Xing Xie. 2023.
\newblock Put your voice on stage: Personalized headline generation for news articles.
\newblock \emph{ACM Trans. Knowl. Discov. Data}, 18(3):20.

\bibitem[{Ao et~al.(2021)Ao, Wang, Luo, Qiao, He, and Xie}]{ao2021pens}
Xiang Ao, Xiting Wang, Ling Luo, Ying Qiao, Qing He, and Xing Xie. 2021.
\newblock {PENS}: A dataset and generic framework for personalized news headline generation.
\newblock In \emph{Proceedings of the 59th Annual Meeting of the Association for Computational Linguistics and the 11th International Joint Conference on Natural Language Processing (Volume 1: Long Papers)}, pages 82--92. Association for Computational Linguistics.

\bibitem[{Bartz et~al.(2008)Bartz, Barr, and Aijaz}]{bartz2008natural}
Kevin Bartz, Cory Barr, and Adil Aijaz. 2008.
\newblock Natural language generation for sponsored-search advertisements.
\newblock In \emph{Proceedings of the 9th ACM Conference on Electronic Commerce}, page 1–9.

\bibitem[{Chai et~al.(2025)Chai, Ren, Xiao, Yang, Han, Zhang, Chen, Lu, Zhao, Yu et~al.}]{chai2025longer}
Zheng Chai, Qin Ren, Xijun Xiao, Huizhi Yang, Bo~Han, Sijun Zhang, Di~Chen, Hui Lu, Wenlin Zhao, Lele Yu, et~al. 2025.
\newblock Longer: Scaling up long sequence modeling in industrial recommenders.
\newblock \emph{arXiv preprint arXiv:2505.04421}.

\bibitem[{Chen et~al.(2025)Chen, Chen, Zhou, Tao, Ding, Xie, Xie, Li, and Feng}]{chen2025enhancing}
Kedi Chen, Qin Chen, Jie Zhou, Xinqi Tao, Bowen Ding, Jingwen Xie, Mingchen Xie, Peilong Li, and Zheng Feng. 2025.
\newblock Enhancing uncertainty modeling with semantic graph for hallucination detection.
\newblock In \emph{Proceedings of the AAAI Conference on Artificial Intelligence}, volume~39, pages 23586--23594.

\bibitem[{Chin-Yew(2004)}]{chin2004rouge}
Lin Chin-Yew. 2004.
\newblock Rouge: A package for automatic evaluation of summaries.
\newblock In \emph{Proceedings of the Workshop on Text Summarization Branches Out, 2004}.

\bibitem[{Dai et~al.(2024)Dai, Pan, Sun, Ji, Xu, Liu, Wang, and Yang}]{dai2024saferlhf}
Josef Dai, Xuehai Pan, Ruiyang Sun, Jiaming Ji, Xinbo Xu, Mickel Liu, Yizhou Wang, and Yaodong Yang. 2024.
\newblock Safe rlhf: Safe reinforcement learning from human feedback.
\newblock In \emph{The Twelfth International Conference on Learning Representations}.

\bibitem[{DeepSeek-AI(2025)}]{deepseekai2025deepseekv3technicalreport}
DeepSeek-AI. 2025.
\newblock Deepseek-v3 technical report.
\newblock \emph{arXiv preprint arXiv:2412.19437}.

\bibitem[{Fan et~al.(2018)Fan, Lewis, and Dauphin}]{fan2018hierarchical}
Angela Fan, Mike Lewis, and Yann Dauphin. 2018.
\newblock Hierarchical neural story generation.
\newblock In \emph{Proceedings of the 56th Annual Meeting of the Association for Computational Linguistics (Volume 1: Long Papers)}, pages 889--898.

\bibitem[{Fujita et~al.(2010)Fujita, Ikushima, Sato, Kamite, Ishiyama, and Tamachi}]{fujita2010automatic}
Atsushi Fujita, Katsuhiro Ikushima, Satoshi Sato, Ryo Kamite, Ko~Ishiyama, and Osamu Tamachi. 2010.
\newblock Automatic generation of listing ads by reusing promotional texts.
\newblock In \emph{Proceedings of the 12th International Conference on Electronic Commerce: Roadmap for the Future of Electronic Business}, page 179–188.

\bibitem[{Holtzman et~al.(2020)Holtzman, Buys, Du, Forbes, and Choi}]{holtzman2020curiouscaseneuraltext}
Ari Holtzman, Jan Buys, Li~Du, Maxwell Forbes, and Yejin Choi. 2020.
\newblock The curious case of neural text degeneration.
\newblock \emph{arXiv preprint arXiv:1904.09751}.

\bibitem[{Huang et~al.(2025)Huang, Chen, Cao, Yang, Qi, Zhu, Han, Liu, Liu, Yao et~al.}]{huang2025towards}
Yanhua Huang, Yuqi Chen, Xiong Cao, Rui Yang, Mingliang Qi, Yinghao Zhu, Qingchang Han, Yaowei Liu, Zhaoyu Liu, Xuefeng Yao, et~al. 2025.
\newblock Towards large-scale generative ranking.
\newblock \emph{arXiv preprint arXiv:2505.04180}.

\bibitem[{Kanungo et~al.(2021)Kanungo, Negi, and Rajan}]{kanungo2021ad}
Yashal~Shakti Kanungo, Sumit Negi, and Aruna Rajan. 2021.
\newblock Ad headline generation using self-critical masked language model.
\newblock In \emph{Proceedings of the 2021 Conference of the North American Chapter of the Association for Computational Linguistics: Human Language Technologies: Industry Papers}.

\bibitem[{Kirk et~al.(2024)Kirk, Mediratta, Nalmpantis, Luketina, Hambro, Grefenstette, and Raileanu}]{kirk2024understandingeffectsrlhfllm}
Robert Kirk, Ishita Mediratta, Christoforos Nalmpantis, Jelena Luketina, Eric Hambro, Edward Grefenstette, and Roberta Raileanu. 2024.
\newblock Understanding the effects of rlhf on llm generalisation and diversity.
\newblock \emph{arXiv preprint arXiv:2310.06452}.

\bibitem[{Lau et~al.(2024)Lau, Hu, Liu, Chen, Ng, and Low}]{lau2024dipperdiversitypromptsproducing}
Gregory Kang~Ruey Lau, Wenyang Hu, Diwen Liu, Jizhuo Chen, See-Kiong Ng, and Bryan Kian~Hsiang Low. 2024.
\newblock Dipper: Diversity in prompts for producing large language model ensembles in reasoning tasks.
\newblock \emph{arXiv preprint arXiv:2412.15238}.

\bibitem[{Li et~al.(2015)Li, Galley, Brockett, Gao, and Dolan}]{li2015diversity}
Jiwei Li, Michel Galley, Chris Brockett, Jianfeng Gao, and Bill Dolan. 2015.
\newblock A diversity-promoting objective function for neural conversation models.
\newblock \emph{arXiv preprint arXiv:1510.03055}.

\bibitem[{Lian et~al.(2025)Lian, Ao, Liu, Liu, and He}]{lian2025panoramic}
Junhong Lian, Xiang Ao, Xinyu Liu, Yang Liu, and Qing He. 2025.
\newblock Panoramic interests: Stylistic-content aware personalized headline generation.
\newblock In \emph{Companion Proceedings of the ACM on Web Conference 2025}, page 1109–1112.

\bibitem[{Liu et~al.(2024)Liu, Feng, Xue, Wang, Wu, Lu, Zhao, Deng, Zhang, Ruan et~al.}]{liu2024deepseek}
Aixin Liu, Bei Feng, Bing Xue, Bingxuan Wang, Bochao Wu, Chengda Lu, Chenggang Zhao, Chengqi Deng, Chenyu Zhang, Chong Ruan, et~al. 2024.
\newblock Deepseek-v3 technical report.
\newblock \emph{arXiv preprint arXiv:2412.19437}.

\bibitem[{Mai and Carson-Berndsen(2024)}]{mai2024linguisticdiversity}
Long Mai and Julie Carson-Berndsen. 2024.
\newblock Improving linguistic diversity of large language models with possibility exploration fine-tuning.
\newblock \emph{arXiv preprint arXiv:2412.03343}.

\bibitem[{Naveed et~al.(2023)Naveed, Khan, Qiu, Saqib, Anwar, Usman, Akhtar, Barnes, and Mian}]{naveed2023comprehensive}
Humza Naveed, Asad~Ullah Khan, Shi Qiu, Muhammad Saqib, Saeed Anwar, Muhammad Usman, Naveed Akhtar, Nick Barnes, and Ajmal Mian. 2023.
\newblock A comprehensive overview of large language models.
\newblock \emph{arXiv preprint arXiv:2307.06435}.

\bibitem[{OpenAI(2024)}]{openai2024gpt4ocard}
OpenAI. 2024.
\newblock Gpt-4o system card.
\newblock \emph{arXiv preprint arXiv:2410.21276}.

\bibitem[{Ouyang et~al.(2022)Ouyang, Wu, Jiang, Almeida, Wainwright, Mishkin, Zhang, Agarwal, Slama, Ray, Schulman, Hilton, Kelton, Miller, Simens, Askell, Welinder, Christiano, Leike, and Lowe}]{ouyang2022traininglanguagemodelsfollow}
Long Ouyang, Jeffrey Wu, Xu~Jiang, Diogo Almeida, Carroll Wainwright, Pamela Mishkin, Chong Zhang, Sandhini Agarwal, Katarina Slama, Alex Ray, John Schulman, Jacob Hilton, Fraser Kelton, Luke Miller, Maddie Simens, Amanda Askell, Peter Welinder, Paul~F Christiano, Jan Leike, and Ryan Lowe. 2022.
\newblock Training language models to follow instructions with human feedback.
\newblock In \emph{Advances in Neural Information Processing Systems}, volume~35, pages 27730--27744.

\bibitem[{Papineni et~al.(2002)Papineni, Roukos, Ward, and Zhu}]{papineni2002bleu}
Kishore Papineni, Salim Roukos, Todd Ward, and Wei-Jing Zhu. 2002.
\newblock Bleu: a method for automatic evaluation of machine translation.
\newblock In \emph{Proceedings of the 40th annual meeting of the Association for Computational Linguistics}, pages 311--318.

\bibitem[{Qwen(2025)}]{qwen2025qwen25technicalreport}
Qwen. 2025.
\newblock Qwen2.5 technical report.
\newblock \emph{arXiv preprint arXiv:2412.15115}.

\bibitem[{Salton and McGill(1986)}]{slaton1986retrieval}
Gerard Salton and Michael~J. McGill. 1986.
\newblock \emph{Introduction to Modern Information Retrieval}.
\newblock McGraw-Hill, Inc., USA.

\bibitem[{Shao et~al.(2024)Shao, Wang, Zhu, Xu, Song, Bi, Zhang, Zhang, Li, Wu et~al.}]{shao2024deepseekmath}
Zhihong Shao, Peiyi Wang, Qihao Zhu, Runxin Xu, Junxiao Song, Xiao Bi, Haowei Zhang, Mingchuan Zhang, YK~Li, Y~Wu, et~al. 2024.
\newblock Deepseekmath: Pushing the limits of mathematical reasoning in open language models.
\newblock \emph{arXiv preprint arXiv:2402.03300}.

\bibitem[{Song et~al.(2023)Song, Chen, Wang, and Shuai}]{song2023general}
Yun-Zhu Song, Yi-Syuan Chen, Lu~Wang, and Hong-Han Shuai. 2023.
\newblock General then personal: Decoupling and pre-training for personalized headline generation.
\newblock \emph{Transactions of the Association for Computational Linguistics}, 11:1588--1607.

\bibitem[{Tan et~al.(2024)Tan, Cheng, Qiu, Shi, Cheng, Chu, Xu, and Qi}]{tan2024offline}
Xiaoyu Tan, Leijun Cheng, Xihe Qiu, Shaojie Shi, Yuan Cheng, Wei Chu, Yinghui Xu, and Yuan Qi. 2024.
\newblock Enhancing personalized headline generation via offline goal-conditioned reinforcement learning with large language models.
\newblock In \emph{Proceedings of the 30th ACM SIGKDD Conference on Knowledge Discovery and Data Mining}, page 5762–5772.

\bibitem[{Tevet and Berant(2021)}]{tevet2021evaluating}
Guy Tevet and Jonathan Berant. 2021.
\newblock Evaluating the evaluation of diversity in natural language generation.
\newblock In \emph{Proceedings of the 16th Conference of the European Chapter of the Association for Computational Linguistics: Main Volume}, pages 326--346, Online. Association for Computational Linguistics.

\bibitem[{Thomaidou et~al.(2013)Thomaidou, Lourentzou, Katsivelis-Perakis, and Vazirgiannis}]{thomaidou2013automated}
Stamatina Thomaidou, Ismini Lourentzou, Panagiotis Katsivelis-Perakis, and Michalis Vazirgiannis. 2013.
\newblock Automated snippet generation for online advertising.
\newblock In \emph{Proceedings of the 22nd ACM International Conference on Information \& Knowledge Management}, page 1841–1844.

\bibitem[{Timmaraju et~al.(2023)Timmaraju, Mashayekhi, Chen, Zeng, Fettes, Cheung, Xiao, Kannadasan, Tripathi, Gahagan et~al.}]{timmaraju2023towards}
Aditya~Srinivas Timmaraju, Mehdi Mashayekhi, Mingliang Chen, Qi~Zeng, Quintin Fettes, Wesley Cheung, Yihan Xiao, Manojkumar~Rangasamy Kannadasan, Pushkar Tripathi, Sean Gahagan, et~al. 2023.
\newblock Towards fairness in personalized ads using impression variance aware reinforcement learning.
\newblock In \emph{Proceedings of the 29th ACM SIGKDD Conference on Knowledge Discovery and Data Mining}, pages 4937--4947.

\bibitem[{Wu et~al.(2023)Wu, Hu, Shi, Dziri, Suhr, Ammanabrolu, Smith, Ostendorf, and Hajishirzi}]{wu2023fine}
Zeqiu Wu, Yushi Hu, Weijia Shi, Nouha Dziri, Alane Suhr, Prithviraj Ammanabrolu, Noah~A Smith, Mari Ostendorf, and Hannaneh Hajishirzi. 2023.
\newblock Fine-grained human feedback gives better rewards for language model training.
\newblock \emph{arXiv preprint arXiv:2306.01693}.

\bibitem[{Xu et~al.(2019)Xu, Wu, Madotto, and Fung}]{xu2019clickbait}
Peng Xu, Chien-Sheng Wu, Andrea Madotto, and Pascale Fung. 2019.
\newblock Clickbait? sensational headline generation with auto-tuned reinforcement learning.
\newblock In \emph{Proceedings of the 2019 Conference on Empirical Methods in Natural Language Processing and the 9th International Joint Conference on Natural Language Processing (EMNLP-IJCNLP)}.

\bibitem[{Yoran et~al.(2023)Yoran, Wolfson, Ram, and Berant}]{yoran2023making}
Ori Yoran, Tomer Wolfson, Ori Ram, and Jonathan Berant. 2023.
\newblock Making retrieval-augmented language models robust to irrelevant context.

\bibitem[{Zhang et~al.(2022)Zhang, Lu, Zhang, Lei, and Wu}]{zhang2022perception}
Kui Zhang, Guangquan Lu, Guixian Zhang, Zhi Lei, and Lijuan Wu. 2022.
\newblock Personalized headline generation with enhanced user interest perception.
\newblock In \emph{Artificial Neural Networks and Machine Learning – ICANN 2022: 31st International Conference on Artificial Neural Networks, Bristol, UK, September 6–9, 2022, Proceedings, Part II}, page 797–809. Springer-Verlag.

\bibitem[{Zhu et~al.(2018)Zhu, Lu, Zheng, Guo, Zhang, Wang, and Yu}]{zhu2018texygen}
Yaoming Zhu, Sidi Lu, Lei Zheng, Jiaxian Guo, Weinan Zhang, Jun Wang, and Yong Yu. 2018.
\newblock Texygen: A benchmarking platform for text generation models.
\newblock In \emph{The 41st international ACM SIGIR conference on research \& development in information retrieval}, pages 1097--1100.

\end{thebibliography}
